\documentclass[conference, 9pt]{IEEEtran}
\ifCLASSINFOpdf
  
\else

\fi

\usepackage[noadjust]{cite}
\usepackage{caption}
\usepackage{url}            
\usepackage{booktabs}       
\usepackage{amsfonts}       
\usepackage{nicefrac}       
\usepackage{microtype}      
\usepackage[cmex10]{amsmath}
\usepackage{amssymb}
\usepackage{amsbsy}
\usepackage{bm}
\usepackage{subfigure}
\usepackage{amsthm}
\usepackage{bbm}
\usepackage{graphicx}
\usepackage{epstopdf}
\usepackage{breqn}
\usepackage{framed}
\usepackage{url}
\usepackage{amsfonts}
\usepackage{mathrsfs}
\usepackage{mathtools}
\DeclareGraphicsExtensions{.pdf}

\DeclareMathOperator*{\st}{s.t.}
\DeclareMathOperator*{\minimize}{minimize}

\DeclareMathOperator*{\support}{supp}

\newtheorem{thm}{Theorem}

\usepackage{spconf}

\begin{document}

%
\title{Efficient estimation of compressible state-space models with application to calcium signal
deconvolution}

\name{Abbas Kazemipour$^{\dagger}$, \qquad Ji Liu$^{\star}$, \qquad Patrick Kanold$^{\star}$, \qquad Min Wu$^{\dagger}$, \qquad Behtash Babadi$^{\dagger}$}

\address{$^{\dagger}$ Department of Electrical and Computer Engineering\\
$^{\star}$ Department of Biology\\
 University of Maryland, College Park, USA\\
\{kaazemi, liuji, pkanold, minwu, behtash\} @umd.edu
}%

\maketitle
\graphicspath{ {./figures/} }

\begin{abstract}
In this paper, we consider linear state-space models with compressible innovations and convergent transition matrices in order to model spatiotemporally sparse transient events. We perform parameter and state estimation using a dynamic compressed sensing framework and develop an efficient solution consisting of two nested Expectation-Maximization (EM) algorithms. Under suitable sparsity assumptions on the innovations, we prove recovery guarantees and derive confidence bounds for the state estimates. We provide simulation studies as well as application to spike deconvolution from calcium imaging data which verify our theoretical results and show significant improvement over existing algorithms.

\end{abstract}


%
\IEEEpeerreviewmaketitle

\section{Introduction}

In many signal processing applications such as estimation of brain activity from MEG time-series \cite{phillips1997meg}, estimation of time-varying networks \cite{kolar2010estimating},  electroencephalogram (EEG) analysis \cite{nunez1995neocortical}, calcium imaging \cite{vogelstein2010fast}, functional magnetic resonance imaging (fMRI) \cite{chang2010time}, and video compression \cite{jung2010motion}, the signals often exhibit abrupt changes which are blurred through convolution with unknown kernels due to the intrinsic measurements constraints. Traditionally, state-space models have been used for estimating the underlying signal given the blurred and noisy observations. Gaussian state-space models in particular are widely used to model smooth state transitions. Under normality assumptions, posterior mean filters and smoothers are optimal estimators, where the analytical solution is given respectively by the Kalman filter and the fixed interval smoother \cite{haykin2008adaptive}.

When applied to observations from abruptly changing states, Gaussian state-space models exhibit poor performance in recovering sharp transitions of the states due to their underlying smoothing property. Although filtering and smoothing recursions can be obtained in principle for non-Gaussian state-space models, exact calculations are no longer possible \cite{fahrmeirstate}. Apart from crude approximations like the extended Kalman filter, several methods have been proposed for state estimation including numerical methods for low-dimensional states \cite{kitagawa1998self}, Monte Carlo filters \cite{kitagawa1998self,hurzeler1998monte}, posterior mode estimation \cite{fruhwirth1994applied,fruhwirth1994data}, and fully Bayesian smoothing using Markov chain Monte Carlo simulation \cite{fahrmeirstate, knorr1999conditional, shephard1997likelihood}. In order to exploit sparsity, several dynamic compressed sensing (CS) techniques, such as the Kalman filtered CS algorithm, have been proposed which typically assume partial information about the sparse support or estimate it in a greedy and online fashion \cite{vaswani2010ls, vaswani2008kalman, carmi2010methods, ziniel2013dynamic, zhan2015time}. However, little is known about the theoretical performance guarantees of these algorithms. 

In this paper, we consider the problem of estimating state dynamics from noisy observations, where the state transitions are governed by autoregressive models with compressible innovations. Motivated by the theory of CS, we employ an objective function formed by the $\ell_1$-norm of the state innovations \cite{ba2012exact}. Unlike the traditional compressed sensing setting, the sparsity is associated with the dynamics and not the states themselves. In the absence of observation noise, the CS recovery guarantees are shown to extend to this problem \cite{ba2012exact}. However, in a realistic setting in presence of observation noise, it is unclear how the CS recovery guarantees generalize to this estimation problem.

We will present stability guarantees for this estimator under a convergent state transition matrix, which confirm that the CS recovery guarantees can be extended to this problem. The corresponding optimization problem in its Lagrangian form is akin to the MAP estimator of the states in a linear state-space model where the innovations are Laplace distributed. This allows us to integrate methods from Expectation-Maximization (EM) theory and Gaussian state-space estimation to derive efficient algorithms for the estimation of states as well as the state transition matrix, which is usually unknown in practice. To this end, we construct two nested EM algorithms in order to jointly estimate the states and the transition matrix. The outer EM algorithm for state estimation is akin to the fixed interval smoother, and the inner EM algorithm uses the state estimates to update the state transition matrix \cite{shumway1982approach}. The resulting EM algorithm is recursive in time, which makes the computational complexity of our method scale linearly with temporal dimension of the problem. This provides an advantage over existing methods based on convex optimization, which typically scale super-linearly with the temporal dimension. Finally, we provide simulation results which reveal that the sparse estimates of the compressible state-space models significantly outperform the traditional basis pursuit estimator. We further apply our estimator to two-photon imaging data for deconvolution of spikes from calcium traces, which confirms the superior performance of our estimator. 

The rest of this paper is organized as follows. In Section \ref{tv:formulation}, we introduce our notation and describe the problem formulation. In Section \ref{sec:tv_theory}, we present our main theoretical result and develop a fast estimator using two nested EM algorithms. We provide simulation studies and application to two-photon imaging data in Section \ref{sec:tv_sim}, followed by concluding remarks in Section \ref{sec:tv_conc}.

\section{Notation and Problem Formulation} \label{tv:formulation}

Throughout the paper we use bold lower and upper case letters for denoting vectors and matrices, respectively. We denote the support of a vector $\mathbf{x}_t \in \mathbb{R}^p$ by $\support(\mathbf{x}_t)$ and its $j$th element by $(\mathbf{x}_{t})_j$. 

We use the notation $[p]:=\{1,2,\cdots,p\}$, and $\mathbf{u}_s$ to denote the best $s$-term approximation to $\mathbf{u}$ in the $\ell_1$-sense. A vector $\mathbf{u}$ of length $p$ is called $s$-sparse (resp. $(s,\xi)$--compressible), if is has $s$ non-zero elements (resp. if $\|\mathbf{u}-\mathbf{u}_s\|_1 \sim \mathcal{O}(s^{\frac{1}{2} - \frac{1}{\xi}})$ for some $\xi \in (0,1)$). We assume the state innovations to be sparse (resp. compressible), i.e. $\mathbf{x}_t-\theta \mathbf{x}_{t-1}$ is $s_t$-sparse (resp. $(s_t,\xi)$--compressible) with $s_1 \gg s_t$ for $t \in [T]\backslash\{1\}$. In the compressive regime that we are interested in, $s_t < n_t \ll p$. For simplicity of notation, we let $\mathbf{x}_0$ to be the all-zero vector in $\mathbb{R}^p$. For an arbitrary set $\mathcal{M} \subset [p]$, $(\mathbf{x}_{t})_{\mathcal{M}}$ denotes the vector $\mathbf{x}_t$ restricted to $\mathcal{M}$, i.e. all the components outside $\mathcal{M}$ set to zero. Given a sparsity level $s$ and a vector $\mathbf{x}$, we denote the set of its $s$ largest magnitude entries by $S$, and its best $s$-term approximation error by $\sigma_s(\mathbf{x}): = \|\mathbf{x}-\mathbf{x}_s\|_1$. 

We consider a linear state-space model given by
\begin{equation}
\label{eq:tv_lap_state_space}
\begin{array}{ll}
\mathbf{x}_t = \theta \mathbf{x}_{t-1}+ \mathbf{w}_t,\\
\mathbf{y}_t = \mathbf{A}_t \mathbf{x}_t + \mathbf{v}_t, &\mathbf{v}_t\sim \mathcal{N}(\mathbf{0},\sigma^2 I)
\end{array},
\end{equation}
where $(\mathbf{x}_t)_{t=1}^{T} \in \mathbb{R}^p$ denote states to be estimated, $\theta$ is the state transition parameter satisfying $|\theta|<1$, $\mathbf{w}_t \in \mathbb{R}^p$ is the innovation sequence, $(\mathbf{y}_t)_{t=1}^T \in \mathbb{R}^{n_t}$ are the linear observations, $\mathbf{A}_t \in \mathbb{R}^{n_t \times p}$ denotes the measurement matrix, and $\mathbf{e}_t \in \mathbb{R}^{n_t}$ denotes the Gaussian measurement noise of known covariance matrix $\sigma^2 \mathbf{I}$. We assume that the innovation $\mathbf{w}_t$ is $(s_t, \xi)$-compressible, and we call this model a \emph{compressible} state-space model to highlight this fact.

For a matrix $\mathbf{A}$, we denote restriction of $\mathbf{A}$ to its first $n$ rows by $(\mathbf{A})_n$ .  We say that the matrix $\mathbf{A} \in \mathbb{R}^{n \times p}$ satisfies the restricted isometry property (RIP) of order $s$, if for all $s$-sparse $\mathbf{x}\in \mathbb{R}^p$, we have
\begin{equation}
\label{eq:tv_rip}
(1-\delta_s) \|\mathbf{x}\|_2^2 \leq \|\mathbf{A}\mathbf{x}\|_2^2 \leq (1+\delta_s)\|\mathbf{x}\|_2^2,
\end{equation}
where $\delta_s \in (0,1)$ is the smallest constant for which Eq. (\ref{eq:tv_rip}) holds \cite{candes2008introduction}. In order to avoid prohibitive storage we assume the rows of $\mathbf{A}_t$ are a subset of rows of $\mathbf{A}_1$, i.e. $\mathbf{A}_t = (\mathbf{A}_{1})_{n_t}$, and define $\tilde{\mathbf{A}}_t = \sqrt{\frac{n_1}{n_t}}\mathbf{A}_t$. In order to promote sparsity of the state dynamics, we consider the dynamic $\ell_1$-regularization (dynamic CS from now on) problem given by
\begin{equation}
\label{eq:tv_prob_def_primal}
\minimize_{\mathbf{x}_1,\mathbf{x}_2,\cdots,\mathbf{x}_T, \theta} \quad \sum_{t=1}^T \frac{\|\mathbf{x}_t-\theta \mathbf{x}_{t-1}\|_1}{\sqrt{s_t}} \;\;\st \;\; \|\mathbf{y}_t-\mathbf{A}_t\mathbf{x}_t\|_2 \leq \sqrt{\frac{n_t}{n_1}}\epsilon.
\end{equation}
Note that this is a variant of the model used in \cite{ba2012exact}. We also consider the (modified) dual form of (\ref{eq:tv_prob_def_primal}) given by
\begin{equation}
\label{eq:tv_prob_def_dual}
\minimize_{\mathbf{x}_1,\mathbf{x}_2,\cdots,\mathbf{x}_T, \theta} \quad \lambda \sum_{t=1}^T \frac{\|\mathbf{x}_t-\theta \mathbf{x}_{t-1}\|_1}{\sqrt{s_t}} +  \frac{1}{n_t}\frac{\|\mathbf{y}_t-\mathbf{A}_t\mathbf{x}_t\|_2^2}{2\sigma^2}.
\end{equation}
Note that Eq. (\ref{eq:tv_prob_def_dual}) is equivalent to the MAP estimator of the states in (\ref{eq:tv_lap_state_space}) if the innovations are given by i.i.d Laplace random variables with parameter $\lambda$. We will next describe the main theoretical results of our paper for stable recovery of the dynamic CS problem.

\section{Theoretical Results and Algorithm Development} \label{sec:tv_theory}

In this section, we state the main theoretical result of our paper regarding the stability of the estimator in \ref{eq:tv_prob_def_dual} and use the EM theory and state-space estimation, to obtain a fast solution to (\ref{eq:tv_prob_def_dual}), which jointly estimates the states as well as their transitions.

\subsection{Stability Guarantees}

Uniqueness and exact recovery of the sequence $(\mathbf{x}_t)_{t=1}^T$ in the absence of noise was proved in \cite{ba2012exact} for $\theta =1$, by an inductive construction of dual certificates. Our main result on stability of the solution of \ref{eq:tv_prob_def_primal} is given in the following Theorem

\begin{thm}[Stable Recovery in the Presence of Noise]
\label{thm:tv_main}
Let $(\mathbf{x}_t)_{t=1}^T \in \mathbb{R}^p$ be a sequence of states such, $\mathbf{A}_1$ and $\tilde{\mathbf{A}}_t$, $t\geq 2$ satisfy RIP of order ${4s}$ with $\delta_{4s} <1/3$, then for fixed known $\theta$, any solution $(\widehat{\mathbf{x}}_t)_{t=1}^{T}$ to (\ref{eq:tv_prob_def_primal}) satisfies
\begin{equation*}
\label{eq:tv_main_stable}
\resizebox{\columnwidth}{!}{$\displaystyle \frac{1}{T} \sum_{t=1}^T \|\mathbf{x}_t-\widehat{\mathbf{x}}_t\|_2 \le\!\frac{1-\theta^T}{1-\theta}\!\! \left(12.6 \left(1+\frac{\sqrt{n_1}-\sqrt{n_2}}{T\sqrt{n_2}}\right) \epsilon + \frac{3}{T}\sum_{t=1}^T \frac{ \sigma_{s_t}(\mathbf{x}_t-\theta \mathbf{x}_{t-1})}{\sqrt{s_t}}\right)$}.
\end{equation*}
\end{thm}

\noindent \textbf{Remarks:} The first term on the right hand side of Theorem \ref{thm:tv_main} implies that the average reconstruction error of the sequence $(\mathbf{x}_t)_{t=1}^T$ is upper bounded proportional to the noise level $\epsilon$, which implies the stability of the estimate. The second term is a measure of compressibility of the innovation sequence and vanishes when the sparsity condition is exactly met.

\noindent \textit{\textbf{Proof Sketch.}} The proof of Theorem \ref{thm:tv_main} is based on establishing a modified cone and tube constraint for the dynamic CS problem and using the boundedness of the Frobenius operator norm of the inverse differencing operator. A full proof can be found in \cite{kazemipour_tv_paper}.

\subsection{Fast Iterative Solution via the EM Algorithm}

In order to obtain a fast solution, we use two nested instances of the EM algorithm. The full details of the algorithm development are given in \cite{kazemipour_tv_paper}, of which we will present a summary in this paper. The outer EM algorithm is also known as the Iteratively Re-weighted Least Squares (IRLS) method \cite{babadi_IRLS}, which aims at estimating the solution to (\ref{eq:tv_prob_def_dual}) in a recursive fashion and is described by iteratively alternating between the following two steps:

\noindent \textbf{Outer E-Step:} In the $(l+1)$-st iteration, given the observed values $(\mathbf{y}_t)_{t=1}^T$, an estimate $(\mathbf{x}_t^{(l)})_{t=1}^T,\theta^{(l)}$ and a small threshold $\epsilon$ the EM algorithm finds the solution to (\ref{eq:tv_prob_def_dual}) as a recursion of
\begin{align}
\label{eq:tv_prob_def_dual_irls}
\minimize_{\mathbf{x}_1,\mathbf{x}_2,\cdots,\mathbf{x}_T, \theta} \quad \frac{\lambda}{2} & \sum_{j=1}^p\sum_{t=1}^T \frac{\left( (\mathbf{x}_{t})_j-\theta (\mathbf{x}_{t-1})_j\right)^2
+\epsilon^2}{\sqrt{s_t}\sqrt{\left( (\mathbf{x}_{t}^{(l)})_j-{\theta}^{(l)} (\mathbf{x}_{t-1}^{(l)})_j\right)^2+\epsilon^2}}\\
\notag &+ \sum_{t=1}^T \frac{1}{n_t}\frac{\|\mathbf{y}_t-\mathbf{A}_t\mathbf{x}_t^{(l)}\|_2^2}{2\sigma^2}.
\end{align}
\textbf{Outer M-Step:}
Given an estimate $(\mathbf{x}_t^{(l)})_{t=1}^T,\theta^{(l)}$, the maximization step of (\ref{eq:tv_prob_def_dual_irls}) involves another instance of EM algorithm index by $m$ as follows:

\noindent \textbf{Inner E-Step:} The main difference between (\ref{eq:tv_prob_def_dual_irls}) and (\ref{eq:tv_prob_def_dual}) is the quadratic form of (\ref{eq:tv_prob_def_dual_irls}). Given an update $\theta^{(l,m)}$, equation (\ref{eq:tv_prob_def_dual_irls}) can be thought of the MAP solution to the Gaussian state-space model given by
\begin{equation}
\label{eq:tv_dynamic}
\begin{array}{l}
\mathbf{x}_t = \theta^{(l,m)} \mathbf{x}_{t-1}+ \mathbf{w}_t, \\ 
\mathbf{w}_t\sim \mathcal{N}\left(\mathbf{0},{\sf diag}\left\{\frac{ {\sqrt{s_t}\sqrt{\left( (\mathbf{x}_{t}^{(l)})_j-{\theta}^{(l)} (\mathbf{x}_{t-1}^{(l)})_j\right)^2+\epsilon^2}}} {\lambda}\right\}_{j=1}^p \right),\\
\mathbf{y}_t = \mathbf{A}_t \mathbf{x}_t + \mathbf{v}_t, \quad \mathbf{v}_t\sim \mathcal{N}(\mathbf{0},n_t\sigma^2 I).
\end{array}
\end{equation}
The inner E-step involves calculation of
\begin{equation}
\label{eq:tv_inner_E}
\mathbb{E}\left\{ \log p \left((\mathbf{y}_t)_{t=1}^T,(\mathbf{x}_t^{(l,m+1)})_{t=1}^T|\theta \right)\Big|(\mathbf{y}_t)_{t=1}^T,\theta^{(l,m)} \right\},
\end{equation}
which can be done by a fixed interval smoother as a fast solution (\ref{eq:tv_dynamic}). We denote the outputs of the smoother in the IRLS algorithm by
\begin{equation*}
\mathbf{x}^{(l, m+1)}_{{t|T}} = {\mathbb{E}}\left \{\mathbf{x}_t\Big|(\mathbf{y}_t)_{t=1}^T, {\theta}^{(l,m)}\right \},
\end{equation*}
\begin{equation*}\mathbf{\Sigma}^{(l,m+1)}_{t|T} = {\mathbb{E}}\left \{ \mathbf{x}_t \mathbf{x}_t'\Big|(\mathbf{y}_t)_{t=1}^T, {\theta}^{(l,m)}\right \},
\end{equation*}
and
\begin{equation*}
\mathbf{\Sigma}^{(l,m+1)}_{t-1,t|T}=\mathbf{\Sigma}^{(l,m+1)}_{t,t-1|T}={\mathbb{E}}\left \{\mathbf{x}_{t-1} \mathbf{x}_t'\Big|(\mathbf{y}_t)_{t=1}^T,{\theta}^{(l,m)}\right \}.
\end{equation*}
The new estimate $(\mathbf{x}_t^{(l,m+1)})_{t=1}^T$ will then replace the estimate of the smoother, that is,
\begin{equation}
\mathbf{x}_t^{(l,m+1)} \leftarrow \mathbf{x}^{(l,m+1)}_{{t|T}}.
\end{equation}
The first and second moments in (\ref{eq:tv_prob_def_dual_irls}) are also replaced using these estimates.

\noindent \textbf{Inner M-Step:} The inner M-step involves maximizing the estimated expectation in (\ref{eq:tv_inner_E}) with respect to $\theta$. Given the observed values $(\mathbf{y}_t)_{t=1}^T$, and an estimate of unobserved values $(\mathbf{x}_t^{(l,m+1)})_{t=1}^T$ and $\theta^{(l,m)}$ the update is given by \cite{kazemipour_tv_paper}:
\begin{equation}
\label{eq:tv_theta_upd}
\theta^{(l, m+1)} = \frac{\sum \limits_{t,j=1}^{T,p} {  \frac{ (\mathbf{x}_{t-1|T})_j (\mathbf{x}_{t|T})_j+ \left(\mathbf{\Sigma}_{t-1,t|T}^{(l,m+1)}\right)_{(j,j)}}{\sqrt{s_t}\sqrt{\left( (\mathbf{x}_{t}^{(l)})_j-{\theta}^{(l)} (\mathbf{x}_{t-1}^{(l)})_j\right)^2+\epsilon^2}}}}{\sum \limits_{t,j=1}^{T,p} {  \frac{  (\mathbf{x}_{t-1|T})_j^2 + \left(\mathbf{\Sigma}_{t-1|T}^{(l,m+1)}\right)_{(j,j)}}{\sqrt{s_t}\sqrt{\left( (\mathbf{x}_{t}^{(l)})_j-{\theta}^{(l)} (\mathbf{x}_{t-1}^{(l)})_j\right)^2+\epsilon^2}}}}.
\end{equation}
This process is repeated for $M$ iterations of the inner loop and $L$ iterations of the outer loop at which a convergence criterion is met. We then have:
\begin{equation*}
{\theta}^{(l+1)} \leftarrow {\theta}^{(l,M)}, \quad ({\mathbf{x}}_t^{(l+1)})_{t=1}^{T} \leftarrow ({\mathbf{x}}_t^{(l,M)})_{t=1}^{T}.
\end{equation*}
\begin{equation*}
\widehat{\theta} \leftarrow {\theta}^{(L)}, \quad (\widehat{\mathbf{x}}_t)_{t=1}^{T} \leftarrow ({\mathbf{x}}_t^{(L)})_{t=1}^{T}.
\end{equation*}

\section{Simulations and Application to Calcium Imaging Data}
\label{sec:tv_sim}
\subsection{Application to Simulated Data}
In this section, we apply the dynamic CS algorithm to simulated data and compare its performance with basis pursuit \cite{chen2001atomic}. We used $p = 200, T = 200, s_1 =8,s_2 = 4,\epsilon = 10^{-10}$, and $\theta = 0.95$. We choose $\frac{s}{n} = \frac{\sum_{t=1}^Ts_t}{\sum_{t=1}^Tn_t}=\frac{s_t}{n_t}$, which is justified by the choice of $n_t = C s_t \log p$ for satisfying the RIP. Theory of LASSO and in general M-estimators \cite{Negahban}, suggests that a good choice for $\lambda$ is given by ${\lambda}  \geq 2\sqrt{2} \sigma \sqrt{\frac{s\log p}{n}}$. We have tuned the choice of $\lambda$ around its theoretical value by cross validation. Moreover we have used estimated the innovation sequence (spikes) from $\widehat{\mathbf{x}}_t-\widehat{\theta} \widehat{\mathbf{x}}_{t-1}$, by thresholding. The thresholding level was chosen using the $90\%$ confidence bounds, such that for the resulting spikes the lower confidence bound of the peaks are higher than the upper confidence bound of preceding troughs.

\begin{figure}[htb!]
\vspace{-2mm}
\centering     
{\includegraphics[width=70mm]{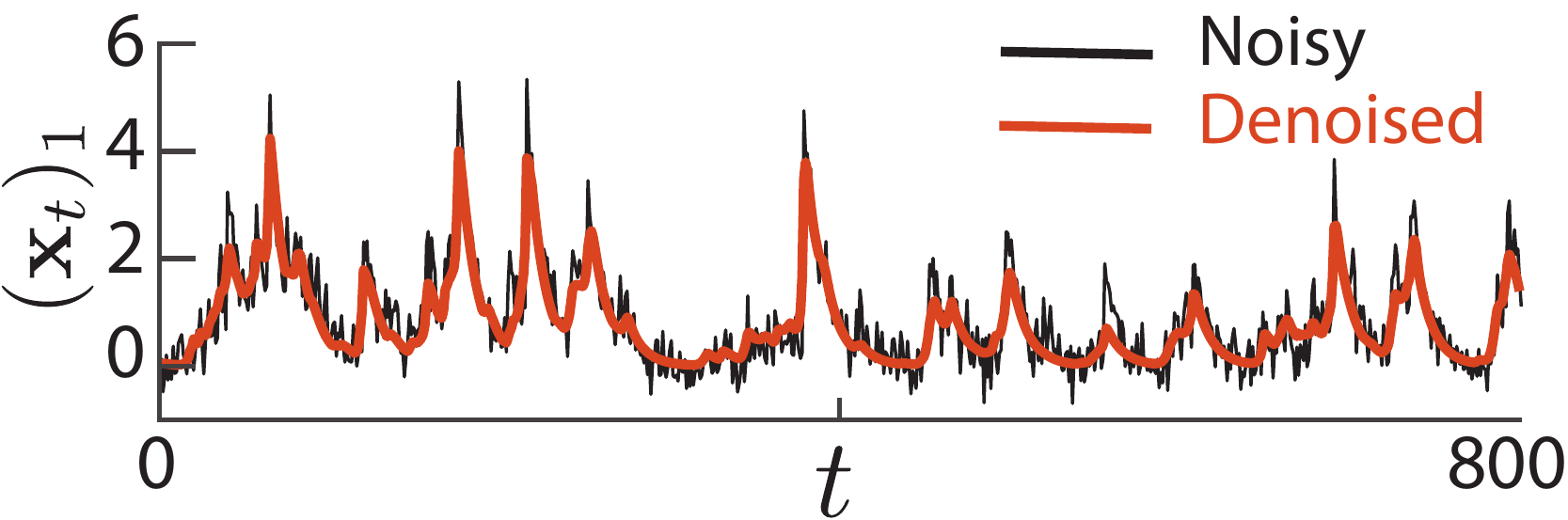}}
\caption{Performance of Dynamic CS on simulated data.}
\label{fig:tv_2ksamples}
\vspace{-2mm}
\end{figure}

\begin{figure}[htb!]
\centering     
{\includegraphics[width=85mm]{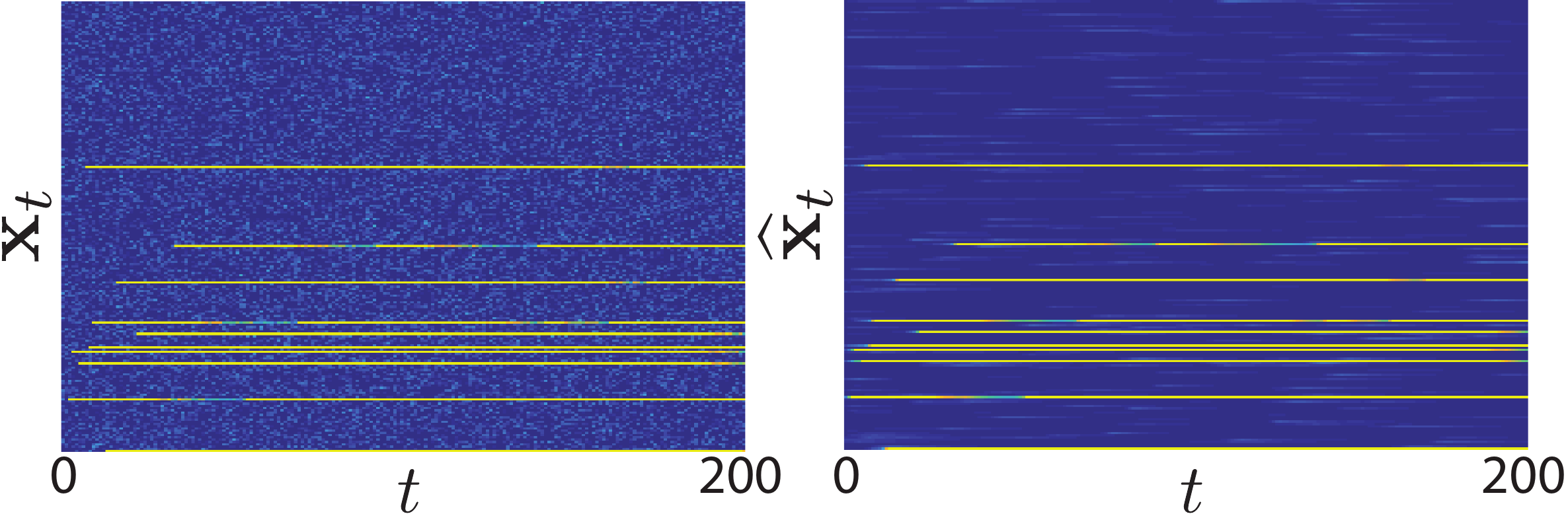}}
\caption{Performance of Dynamic CS: noisy (left) vs. denoised (right) data}
\label{fig:tv_raster_simulated}
\vspace{-4mm}
\end{figure}

Figure \ref{fig:tv_2ksamples} shows $800$ samples of $(\mathbf{x}_t)_1$ (black trace) and its denoised version (red trace) for an SNR value of $5~\text{dB}$. The denoised signal tracks the jumps in the state sequence which significantly denoising the trace. Figure \ref{fig:tv_raster_simulated} shows the simulated and estimated states across time. The estimated states are significantly denoised while the sparsity structure is preserved.

\begin{figure}[htb!]
\centering     
\subfigure[State Dynamics]{\label{fig:tv_ds}\includegraphics[width=40mm]{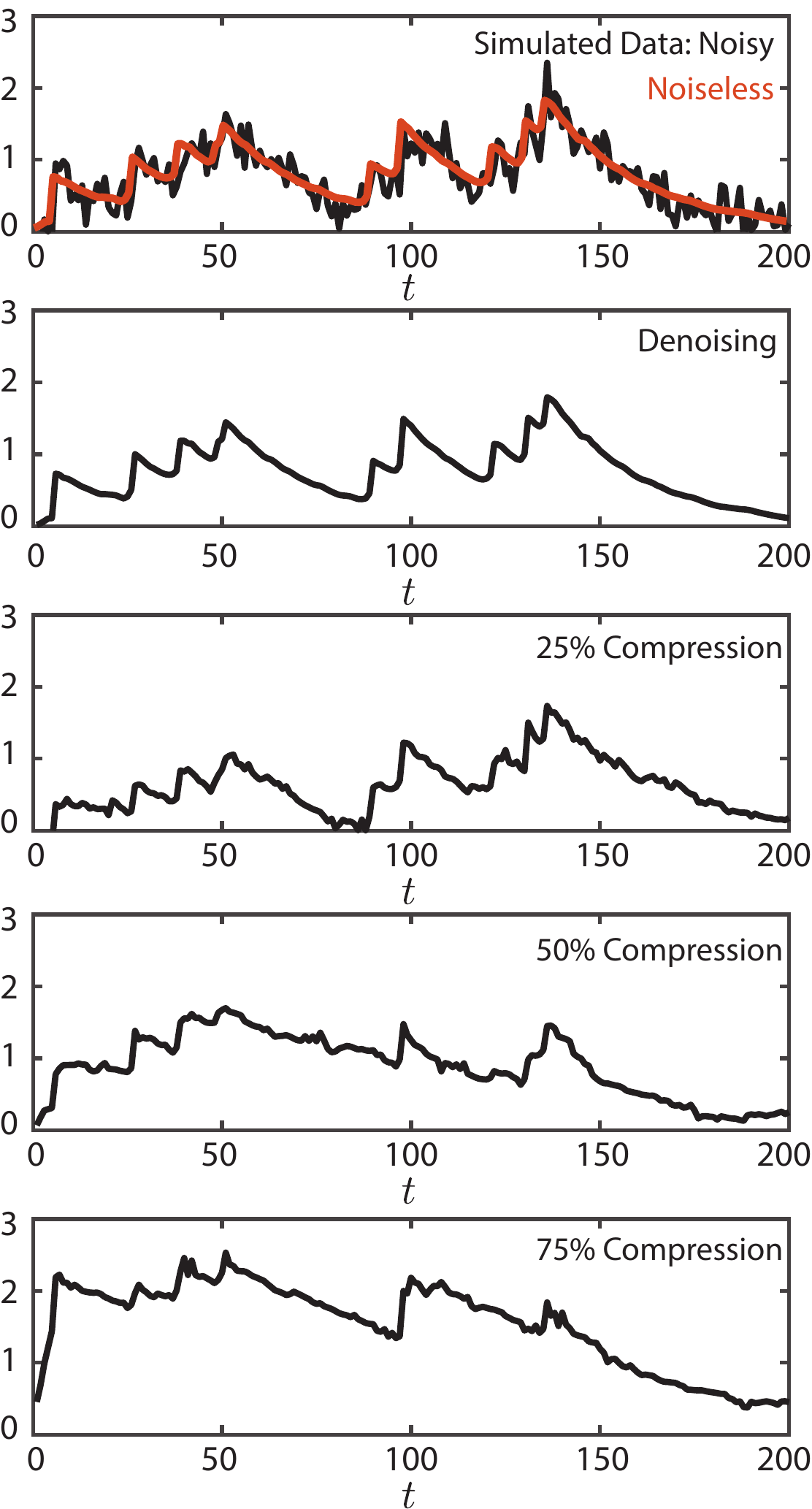}}
\subfigure[Ground-Truth Spikes]{\label{fig:tv_ds_spikes}\includegraphics[width=41.5mm]{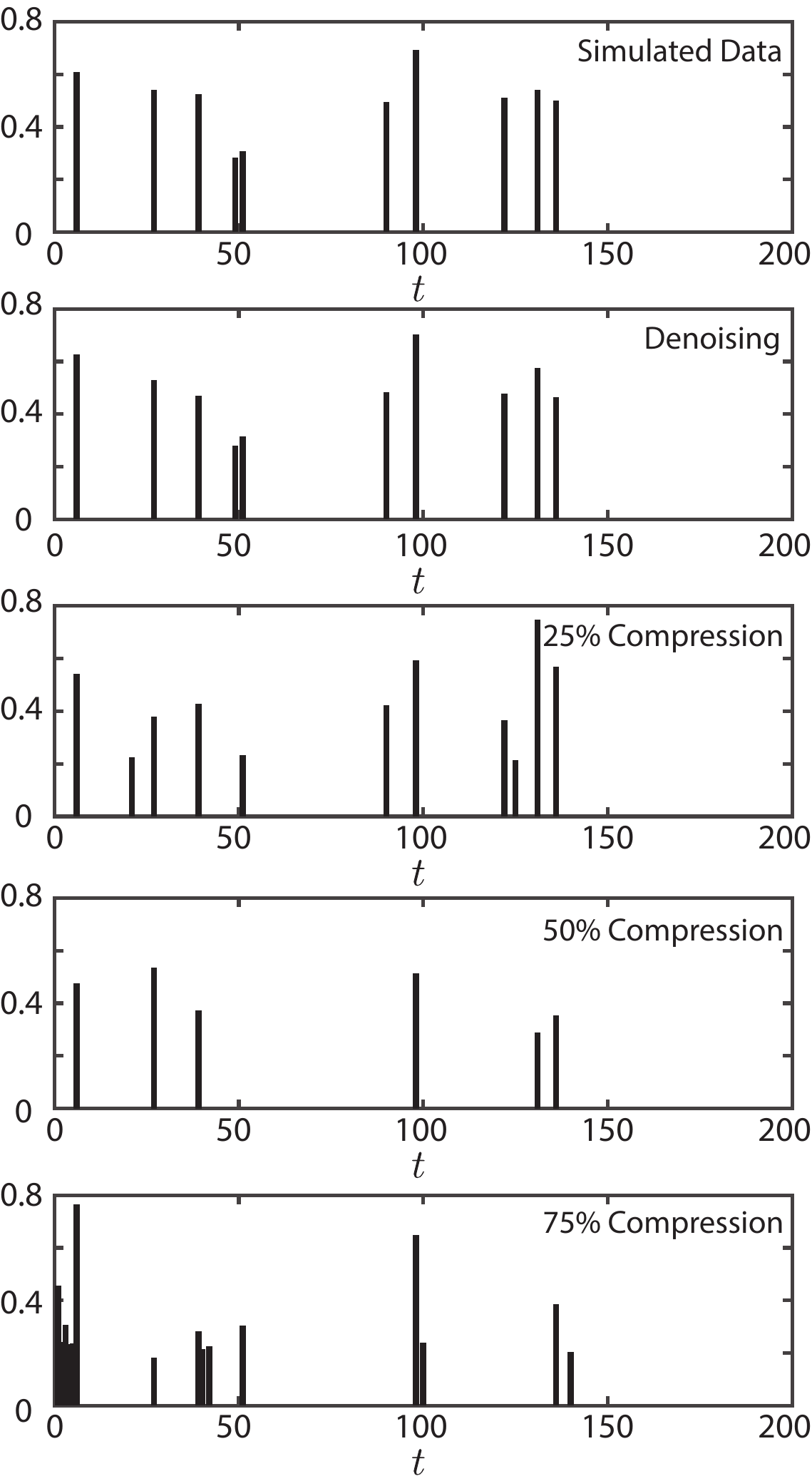}}
\caption{Reconstructed states (left) and spikes (right) using Dynamic CS with varying compression level.}
\vspace{-4mm}
\end{figure}

%
Figures \ref{fig:tv_ds} and \ref{fig:tv_ds_spikes} show respectively the denoised traces and the detected spikes for varying compression levels of $1-n/p = 0, 0.25, 0.5,$ and $0.75$. As the compression level increases, the performance of the algorithm degrades, but strikingly the significant spikes can still be detected at a compression level of $0.75$.

\begin{figure*}[htb!]
\vspace{-3mm}
\centering     
\subfigure[Raw Calcium Imaging Data (Noisy)]{\label{fig:tv_cal_true}\includegraphics[width=43.5mm]{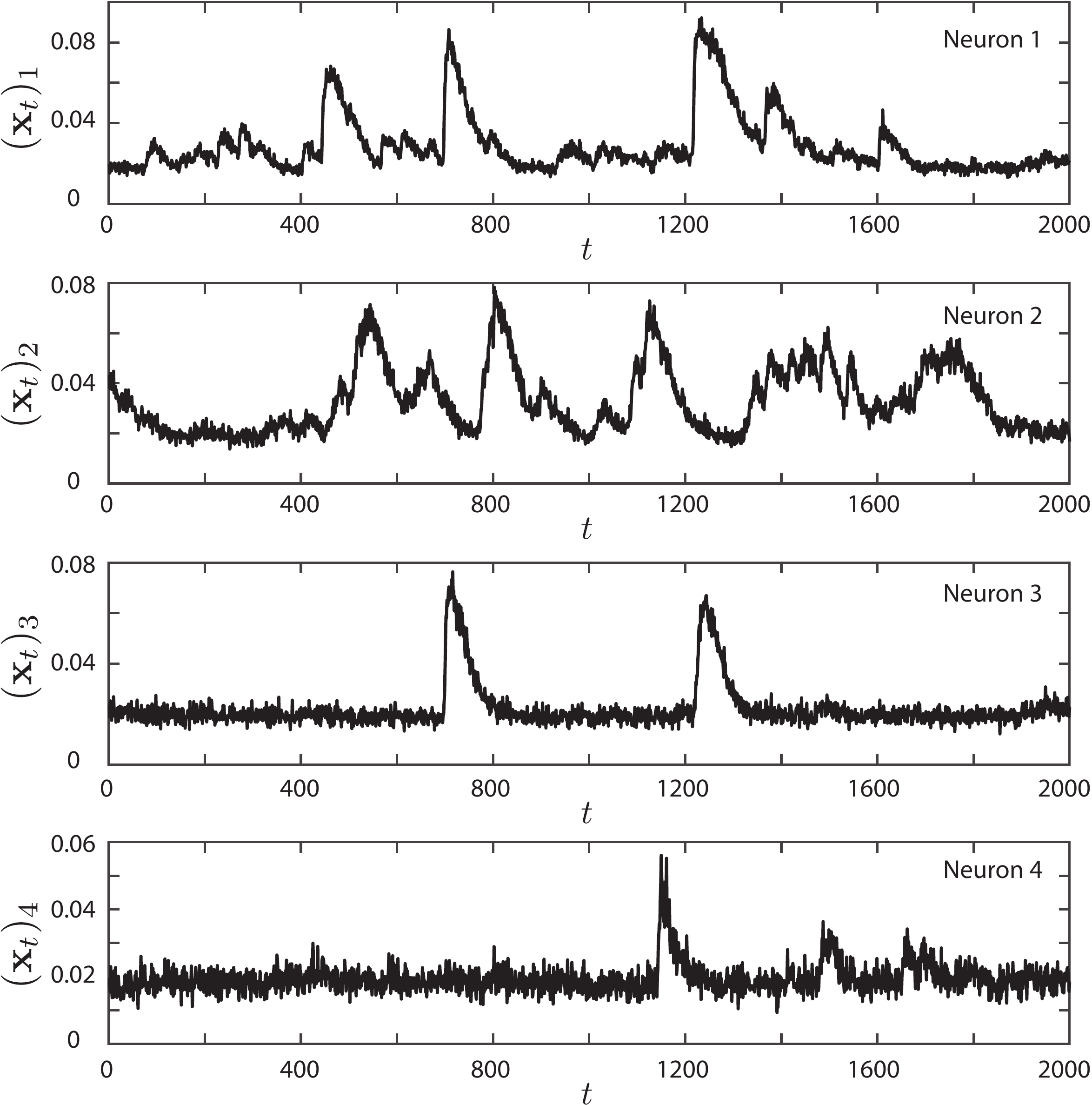}}
\subfigure[Reconstructed States After Denoising]{\label{fig:tv_cal_denoising}\includegraphics[width=43.75mm]{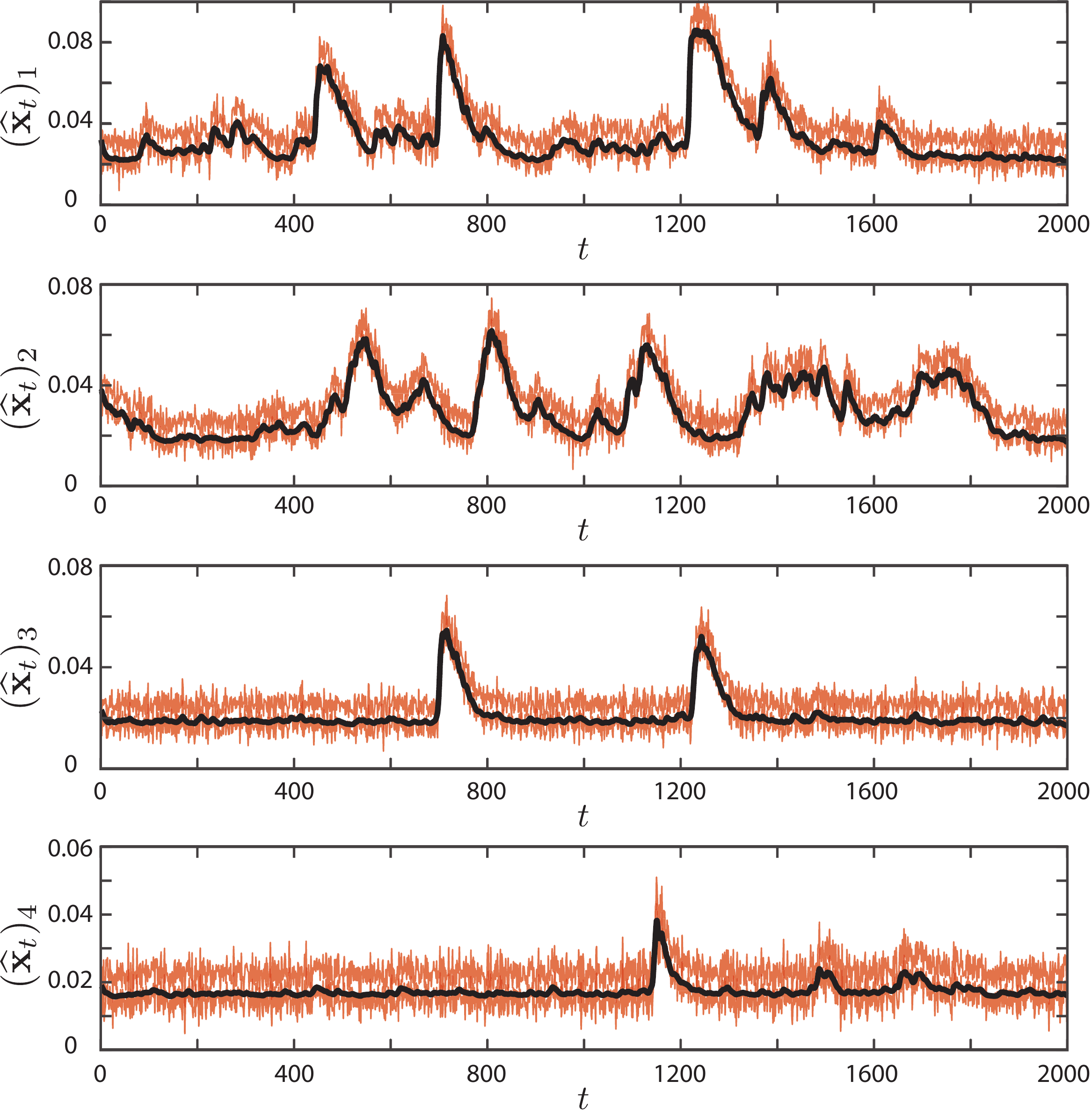}}
\subfigure[Reconstructed States After Denoising and Compression $n/p = 2/3$]{\label{fig:tv_cal_comp}\includegraphics[width=43.75mm]{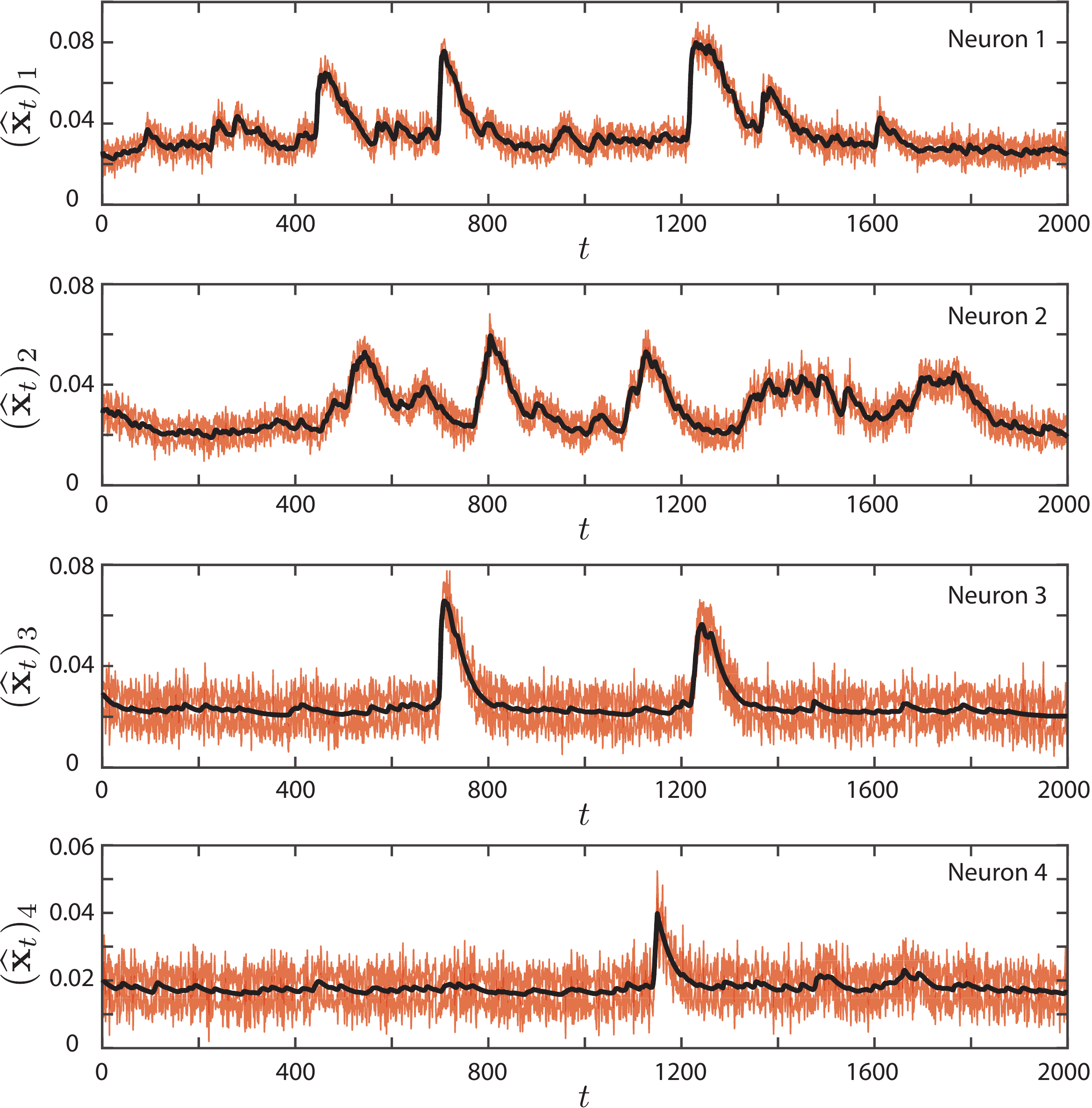}}
\caption{Performance of Dynamic CS on calcium imaging data}
\label{fig:tv_cal}
\vspace{-4mm}
\end{figure*}

\begin{figure*}[htb!]
\centering     
\subfigure[Estimated spikes from constrained f-oopsi algorithm]{\label{fig:tv_spikes_foopsi}\includegraphics[width=43.2mm]{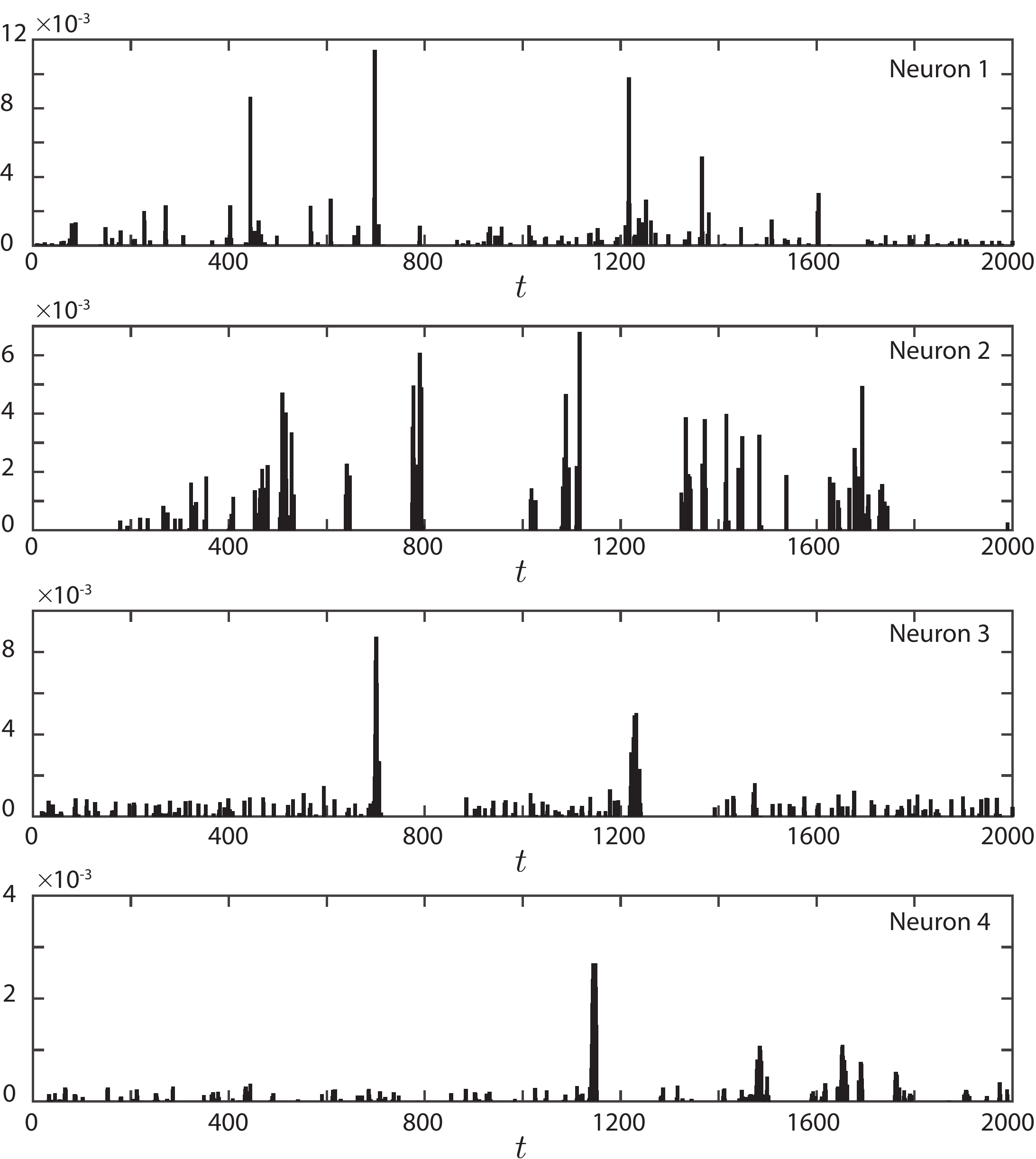}}
\subfigure[Estimated spikes from Dynamic CS after denoising]{\label{fig:tv_spikes_dcs}\includegraphics[width=44mm]{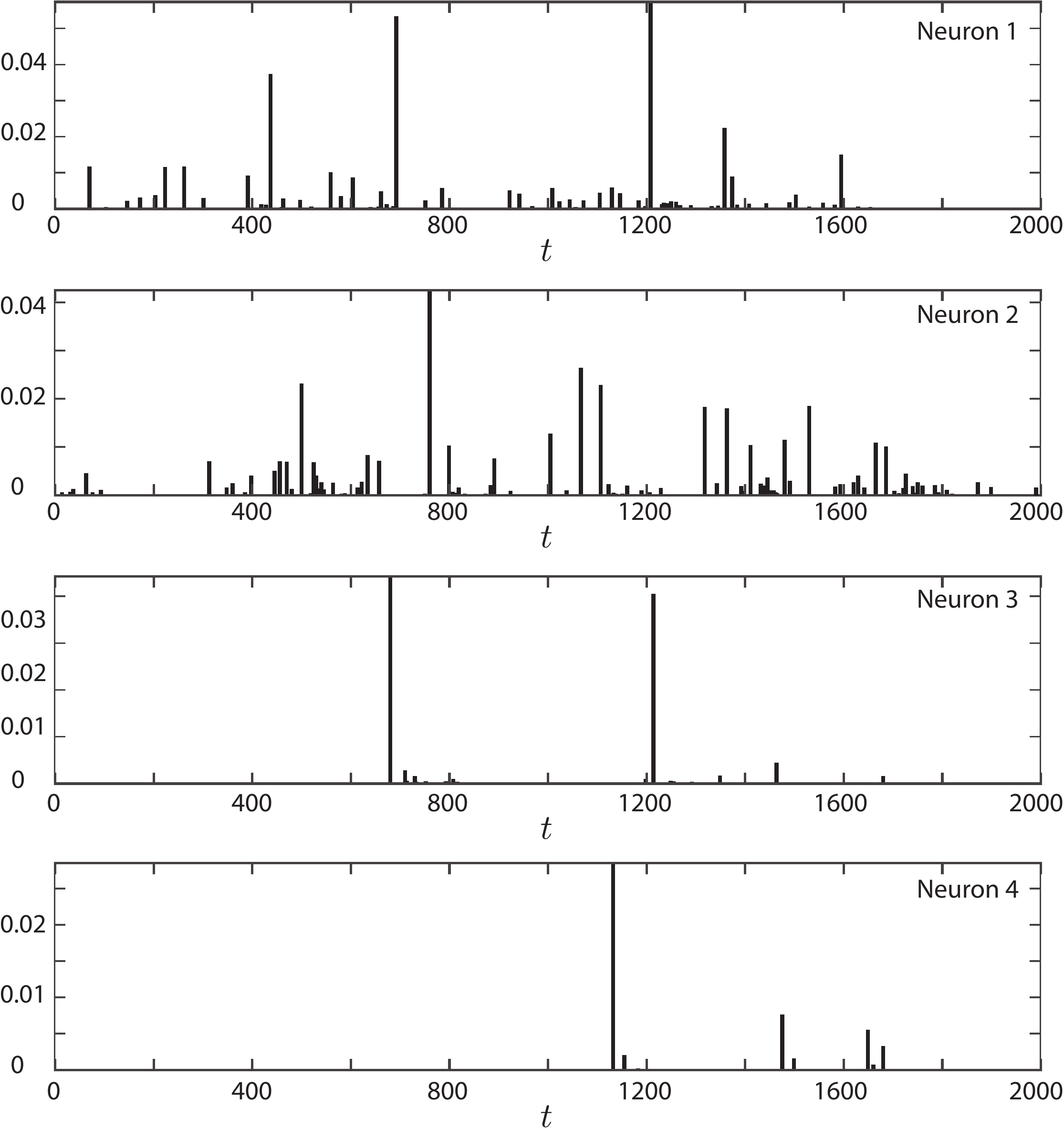}}
\subfigure[Estimated spikes from Dynamic CS after denoising and compression $n/p = 2/3$]{\label{fig:tv_spikes_dcs}\includegraphics[width=44mm]{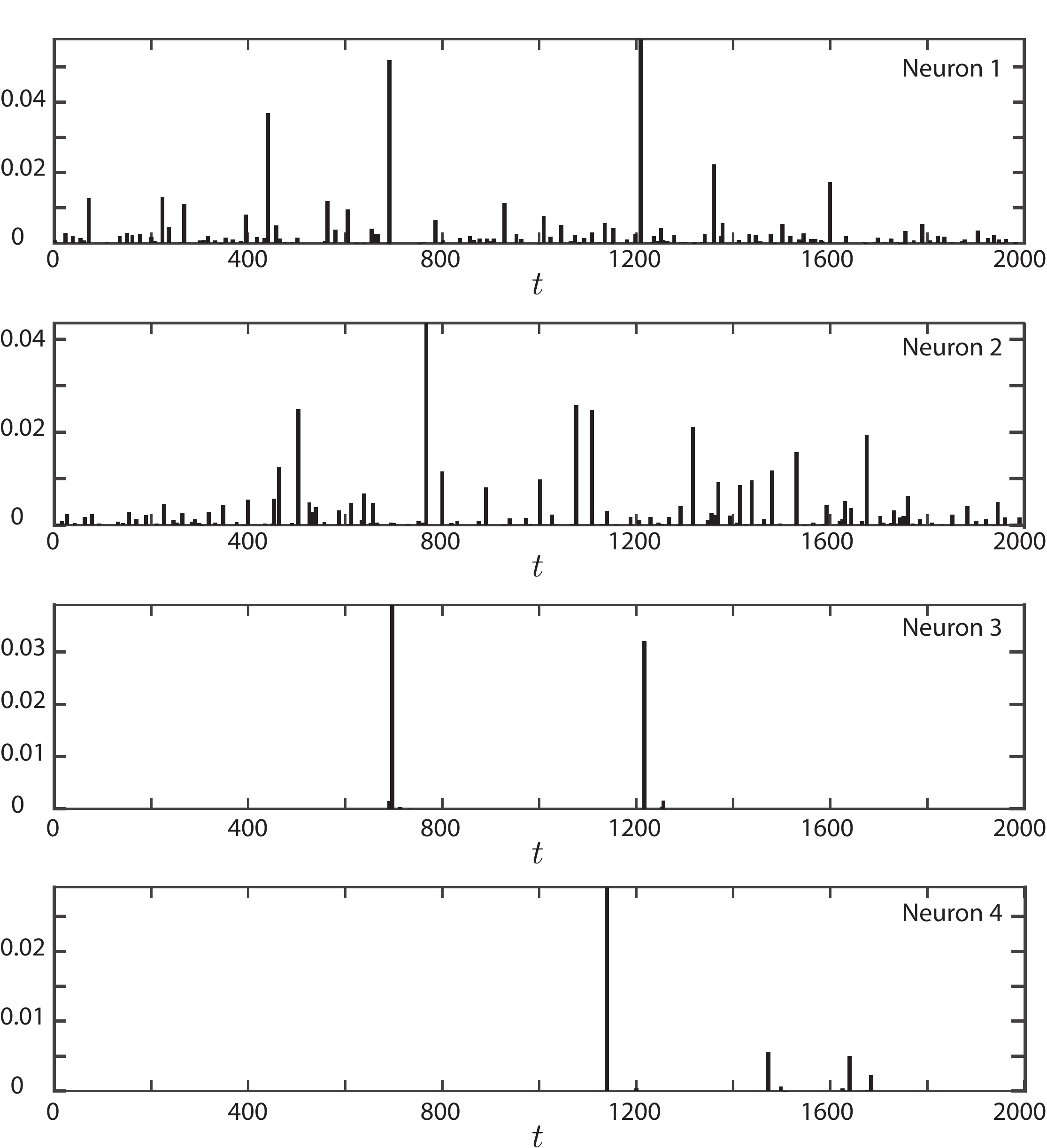}}
\caption{Reconstructed spikes of Dynamic CS from calcium imaging data}
\label{fig:tv_cal_spikes}
\vspace{-4mm}
\end{figure*}

\subsection{Application to Calcium Imaging Data of Neural Spiking Activities}
In this section we apply the dynamic CS algorithm to real data recordings of calcium traces of neuronal activity. Calcium imaging takes advantage of intracellular calcium flux to directly visualize calcium signaling in living neurons. This is done by using Calcium indicators, fluorescent molecules that can respond to the binding of Calcium ions by changing their fluorescence properties and using a fluorescence microscope and a CCD camera to capture record the visual patterns \cite{smetters1999detecting,stosiek2003vivo}. The data was recorded from $219$ neurons at a rate of $30$ frames per second for a total time of $22$ minutes from the mouse's auditory cortex using a two-photon microscope. We chose $T=2000$ samples corresponding to $1$ minute for analysis. In order to suppress the neuropil effects, the data were spatially filtered. We chose $p=108$ spatially separated neurons by visual inspection. We estimate the measurement noise variance from an inactive period of spiking activities to be $\sigma^2 = 10^{-5}$ and use a value of $\epsilon = 10^{-10}$. Figure \ref{fig:tv_cal} shows the denoised states for four sample neurons with $90\%$ confidence bounds. The output is significantly denoised while preserving the dynamics of the data.

\begin{figure}[htb!]
\vspace{-3mm}
\centering     
{\includegraphics[width=85mm]{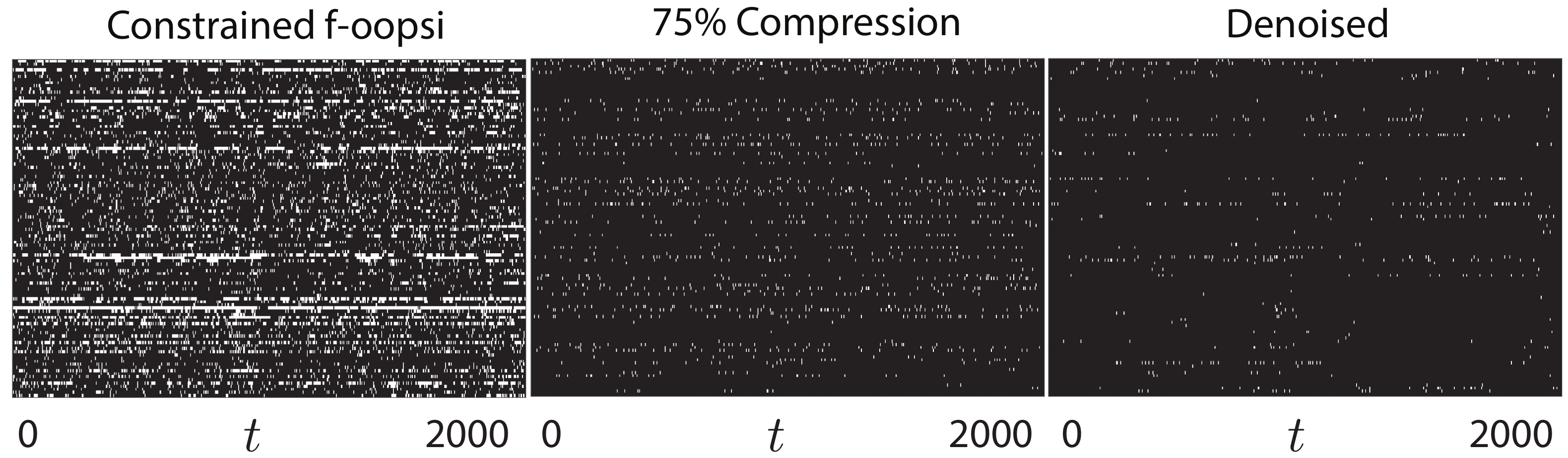}}
\caption{Raster plot of the estimated spikes.}
\label{tv_raster_cal}
\vspace{-6mm}
\end{figure}

Figure \ref{fig:tv_cal_spikes} shows the reconstruction of the spikes in comparison to the constrained f-oopsi algorithm \cite{vogelstein2010fast}, which assumes an inhomogeneous Poisson model for spiking with an exponential approximation. Similar to the simulated data, the thresholding level was chosen using the confidence bounds. Note that the performance of our algorithm remains largely the same when $2/3$ of the observations are used. Figure \ref{tv_raster_cal} shows the corresponding raster plot of the detected spikes. By comparing the performance of f-oopsi to our algorithm, two observations can be made. First, the f-oopsi algorithm outputs a large number of small spikes, whereas our algorithm rejects them. Second, the detected events of f-oopsi are in the form of spike clusters, whereas our algorithm outputs correspondingly separated spikes. This difference in performance is due to the fact that we explicitly model the sparse nature of the spiking activity by going beyond the Gaussian state-space modeling paradigm. In contrast, the constrained f-oopsi algorithm assumes an exponential approximation with a log-barrier to a Poisson model of spiking activities, which results in losing the temporal resolution of the jumps. 
In addition, we are able to form precise confidence bounds for our estimates, whereas the f-oopsi algorithm does not produce statistical confidence bounds. Our thresholding method is based on these confidence bounds which results in a systematic detection and rejection of the spikes. 


\section{Conclusions}
\label{sec:tv_conc}
In this paper, we considered compressible state-space models, where the state innovations are modeled by a sequence of compressible vectors. The traditional results of CS theory do not readily generalize to this cases where the sparsity lies in the dynamics and not the state itself, as the overall linear measurement operator does not satisfy regularity conditions such as the RIP \cite{ba2012exact}. We showed that the guarantees of CS can indeed be extended to the state estimation problem. Hence, using the state-space model, one can infer temporally global information from local measurements.

We also developed a scalable, low-complexity algorithm using two nested EM algorithms for the estimation of the states as well as the transition parameter. We further verified the validity of our theoretical results through simulation studies as well as application to real data recordings of calcium traces of neuronal activity. In addition to scalability and the ability to track the rapid dynamics in the states and in comparison to the widely used spike deconvolution algorithm f-oopsi, our algorithm provides a systematic way to detect the spike events by forming statistical confidence intervals for the state estimates. Our results suggest the possibility of using compressive measurements for reconstruction and denoising of calcium traces, which from a practical point of view, can allow faster data acquisition by undersampling the field of view . We consider joint spike sorting and deconvolution as future work.



\section{Acknowledgments}

This work was supported in part by the National Institutes of Health Award No. R01DC009607 and the National Science Foundation Award No. 1552946.

{
\small
\bibliographystyle{IEEEtran}
\bibliography{TV}
}

\end{document}